\documentclass{article}

\usepackage{microtype}
\usepackage{graphicx}
\usepackage{subfigure}
\usepackage{booktabs} 
\usepackage{amsfonts}
\usepackage{listings}
\usepackage{hyperref}

\newcommand{\fig}[1]{Figure~\ref{#1}}
\newcommand{\tab}[1]{Table~\ref{#1}}

\usepackage[accepted]{icml2023}
\usepackage{amsmath}
\usepackage{amssymb}
\usepackage{mathtools}
\usepackage{amsthm}
\usepackage{multirow}
\usepackage{array}
\usepackage{subcaption}
\usepackage[textwidth=2.5cm]{todonotes}

\usepackage[capitalize,noabbrev]{cleveref}

\theoremstyle{plain}

\theoremstyle{definition}

\theoremstyle{remark}

\icmltitlerunning{Weight subcloning: direct initialization of transformers using larger pretrained ones}

\begin{document}

\newcolumntype{C}{@{}c@{\hspace{0.5em}}}

\twocolumn[
\icmltitle{Weight Subcloning: \\ Direct Initialization of Transformers Using Larger Pretrained Ones}



\icmlsetsymbol{equal}{*}

\begin{icmlauthorlist}
\icmlauthor{Mohammad Samragh}{}
\icmlauthor{Mehrdad Farajtabar}{}
\icmlauthor{Sachin Mehta}{}
\icmlauthor{Raviteja Vemulapalli}{}\\
\icmlauthor{Fartash Faghri}{}
\icmlauthor{Devang Naik}{}
\icmlauthor{Oncel Tuzel}{}
\icmlauthor{Mohammad Rastegari}{}

\icmlauthor{Apple}{}
\end{icmlauthorlist}


\icmlcorrespondingauthor{Mohammad Samragh}{m_samraghrazlighi@apple.com}

\icmlkeywords{Machine Learning, ICML}

\vskip 0.3in
]




\DeclareRobustCommand{\mehrdad}[1]{\todo[color=blue!30,size=\tiny]{Mehrdad: #1}}

\begin{abstract}
Training large transformer models from scratch for a target task requires lots of data and is computationally demanding. The usual practice of transfer learning overcomes this challenge by initializing the model with weights of a pretrained model of the \emph{same size} and specification to increase the convergence and training speed. However, what if no pretrained model of the required size is available? In this paper, we introduce a simple yet effective technique to transfer the knowledge of a pretrained model to smaller variants. Our approach called weight subcloning expedites the training of scaled-down transformers by initializing their weights from larger pretrained models.

Weight subcloning involves an operation on the pretrained model to obtain the equivalent initialized scaled-down model. It consists of two key steps: first, we introduce neuron importance ranking to decrease the embedding dimension per layer in the pretrained model. Then, we remove blocks from the transformer model to match the number of layers in the scaled-down network. The result is a network ready to undergo training, which gains significant improvements in training speed compared to random initialization. For instance, we achieve $4\times$ faster training for vision transformers in image classification and language models designed for next-token prediction.

\end{abstract}

\section{Introduction}




Transformers are models extensively used for various tasks, including language modeling~\citep{radford2019language,dai2019transformer,zhang2023survey} and vision applications~\citep{han2022survey, dosovitskiy2020image}. 
Training transformer models typically requires extensive computing resources and large-scale training data. Pretrained models, developed by the research and industry community, are available to transfer the learned weights for a wide range of applications~\citep{gpt2_hf}.
Nevertheless, in practical deployment, it is often necessary to train a scaled-down version of the pretrained model that better aligns with the available hardware resources. Training a scaled-down model from scratch requires substantial GPU hours especially for models that are inherently complex and hard to train, e.g., large language models (LLMs). 

This paper aims to speed up the training of a scaled-down transformer by using pretrained model weights during initialization. We refer to the pretrained and scaled-down networks as the parent and destination models, respectively. In our setup, the destination network has fewer layers and/or smaller hidden dimensions per layer than the parent model. Our primary question is: can we transfer the knowledge of the parent model to the destination model? In other words, does initializing the destination network with weights derived from the pre-trained parent model lead to improved training speed and potentially better accuracy? 

\begin{figure}[ht]
    \centering
    \begin{subfigure}{}
        \includegraphics[width=0.45\linewidth]{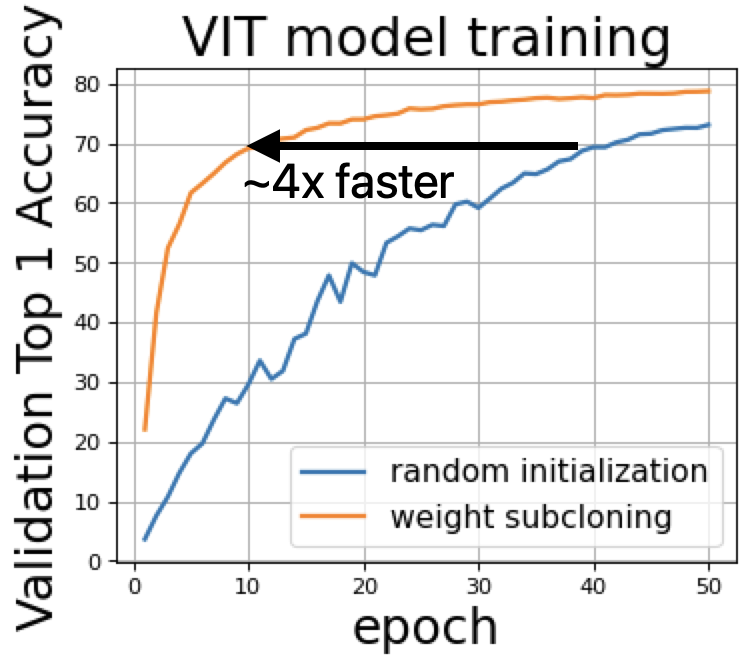}
    \end{subfigure}
    \hfill
    \begin{subfigure}{}
        \includegraphics[width=0.45\linewidth]{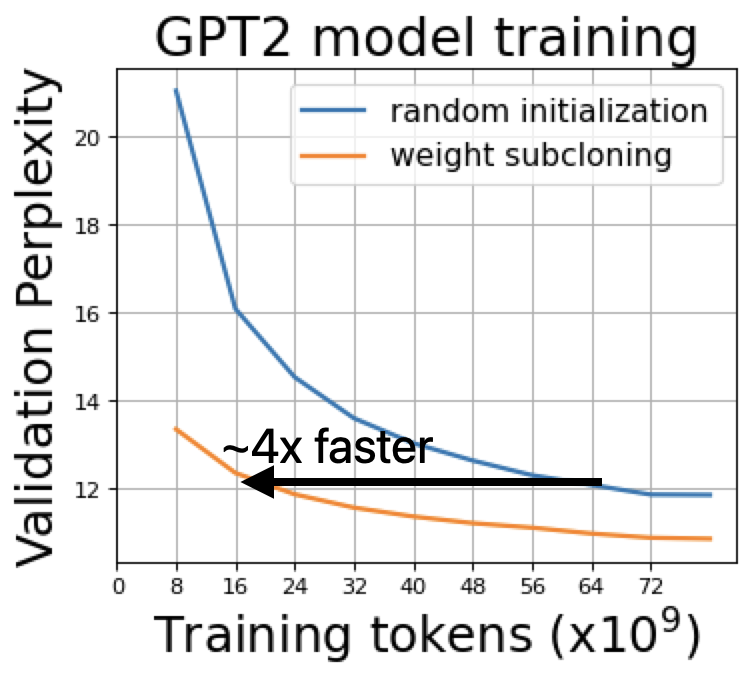}
    \end{subfigure}
    \caption{Validation accuracy and perplexity for destination network training for two tasks. left: image classification with VIT networks~\cite{dosovitskiy2020image}. right: next token prediction with GPT-2 networks~\cite{radford2019language}. Weight subcloning reduces the number of iterations required to achieve an early accuracy/perplexity, and increases the final performance achieved in a limited number of iterations.}
    \label{fig:motivation}
\end{figure}

We refer to such weight transfer process as weight subcloning. The problem of transferring knowledge from a parent model to a destination model has been investigated in different contexts, such as knowledge distillation~\citep{gou2021knowledge, lin2022knowledge, park2022self}, weight sharing (also known as supernet training)~\citep{wang2021attentivenas, cai2019once, yu2020bignas, wang2021alphanet}, and pruning~\citep{blalock2020state, han2015learning,he2017channel}. We will elaborate on each of these research tracks and their relation to our solution in Section~\ref{sec:related}. Nevertheless, our findings demonstrate that weight subcloning is a low-cost yet effective approach for enhancing the training speed of transformer models. In summary, the contributions of this paper are as follows: 


\begin{itemize}
    \item We introduce the concept of weight subcloning for transformer models, which arises when there's a pretrained network already in place, and a smaller destination network is set to undergo training. 
    \item We demonstrate the ability to remove or duplicate transformer blocks within a pretrained transformer network, enabling us to initialize networks with depths either lower or higher than that of the pretrained model.
    \item By uncovering a consistent pattern of neuron importance across transformer layers, we establish the foundation for a re-ordering technique. This method organizes network parameters, consistently ranking them from most important to least important within each layer. Consequently, a destination network with a smaller hidden dimension can selectively utilize the most crucial weights from the parent network.
    \item Our findings illustrate that weight subcloning significantly enhances the training speed for the destination network (See \fig{fig:motivation}). 
\end{itemize}

\section{Related Work}\label{sec:related}
Weight distillation is related to various research pathways. We summarize the most relevant ones in this section.

\subsection{Distillation}

Knowledge distillation is a technique used to create a smaller student model from a larger teacher model, with the goal of reducing model size and computational complexity~\cite{gou2021knowledge, lin2022knowledge, park2022self}. In this approach, the student model is trained to imitate the teacher model, either at the output layer or by replicating intermediate hidden features. This process can be time-consuming because it involves the large teacher model during training. In a closely related work to our paper,~\cite{lin2020weight} propose a method called ``weight distillation''. In weight distillation, in addition to logit alignment, the parameters of the teacher are mapped to the student network using a transformation, which is learned in the presence of both teacher and student networks during training.

In the context of knowledge and weight distillation, we can view the teacher and student models as the parent and destination architectures, respectively, in our scenario. What distinguishes our approach from traditional distillation techniques is our focus on achieving faster training. Instead of actively involving the teacher model in the training process, we directly transfer the weights from the parent model to the destination model. In comparison to weight distillation, we demonstrate that the function used to map parent to destination parameters can be as straightforward as cloning with minor modifications, eliminating the need for additional learning during this parameter transfer. As a direct consequence of this approach, the training loop remains unchanged, making it significantly more adaptable for a wide range of training tasks.

\subsection{Supernet training}

To mitigate the engineering and GPU resource costs associated with architecture selection, a common strategy is weight sharing~\cite{wang2021attentivenas, cai2019once, yu2020bignas, wang2021alphanet}. In this approach, network parameters are consolidated within a supernet, and smaller subnets utilize a subset of the supernet's weights during forward and backward propagation. At test time, the subnets within the supernet can be extracted and deployed, offering various model sizes and accuracy levels. 

Supernet training operates by randomly selecting subnets during each training iteration. Consequently, the effective number of training epochs for each subnet is roughly equivalent to the total number of training epochs divided by the number of subnets visited during training. As a result, supernet training generally demands more training time compared to standard network training. Additionally, supernets may encounter convergence challenges stemming from gradient conflicts (parameter update conflicts) between subnets. Addressing these convergence issues is notably challenging for large-scale models like LLMs. 

The common element of this paper and supernet training is in their problem definition. The parent and destination models of our setting can be considered as the supernet and subnet in the weight sharing literature. However, it's important to note that the majority of publicly available pretrained models have not been subject to the supernet-style training approach. Consequently, simply extracting subnets from these models may not yield the desired accuracy.

We delve into transformer architectures and demonstrate that transformers inherently exhibit supernet-like characteristics, without undergoing supernet-style training. In other words, we establish that transformer weights can serve as effective initialization for derived subnets even when the pre-trained network has not undergone supernet-style training.

\subsection{Pruning}

Pruning is a method used to create a compact model from a larger, pretrained model~\cite{blalock2020state, han2015learning}. Pruning techniques selectively reduce the size of the model by making some weight parameters zero, either in a structured or non-structured way. 

Non-structured pruning, often referred to as weight pruning, may appear somewhat unrelated to our specific problem. This pruning technique aims to reduce the parameter size in a network without altering its architecture. In certain instances of non-structured pruning, such as the lottery tickets hypothesis proposed by~\cite{frankle2018lottery}, the pruned network is even re-initialized in order to match or attain improved end-accuracy, rather than achieving a faster convergence. This distinction further sets these works apart from our specific problem setting.

Structured pruning is particularly relevant to our problem because it modifies the network's architecture, typically by reducing the number of neurons in each layer~\cite{he2017channel}. To be more precise, structured pruning involves extracting a smaller sub-network from a pretrained model and then fine-tuning it. One distinctive feature that sets our approach apart from structured pruning is our neuron ordering method. It can be seen as a solution to an exceptionally constrained pruning problem, where the number and index of preserved neurons must remain consistent across all transformer layers. Ultimately, what distinguishes our work from traditional pruning techniques is our primary emphasis on achieving faster training within a reduced number of epochs, rather than striving for higher accuracy through extended training periods.

\section{Insights}\label{sec:insights}

An established observation in architectures featuring residual connections, as highlighted by He et al.~\cite{he2016deep}, is that individual blocks within residual networks induce only slight changes to the hidden representation. Transformers, a specific type of residual network, also exhibit this characteristic. We hereby refer to this characteristic as the additive residual property. This property asserts that tokens at layer $i$ resemble those at layer $i-1$~\cite{liu2023deja}.

\begin{figure*}[ht]
    \centering
    \includegraphics[width=0.8\textwidth]{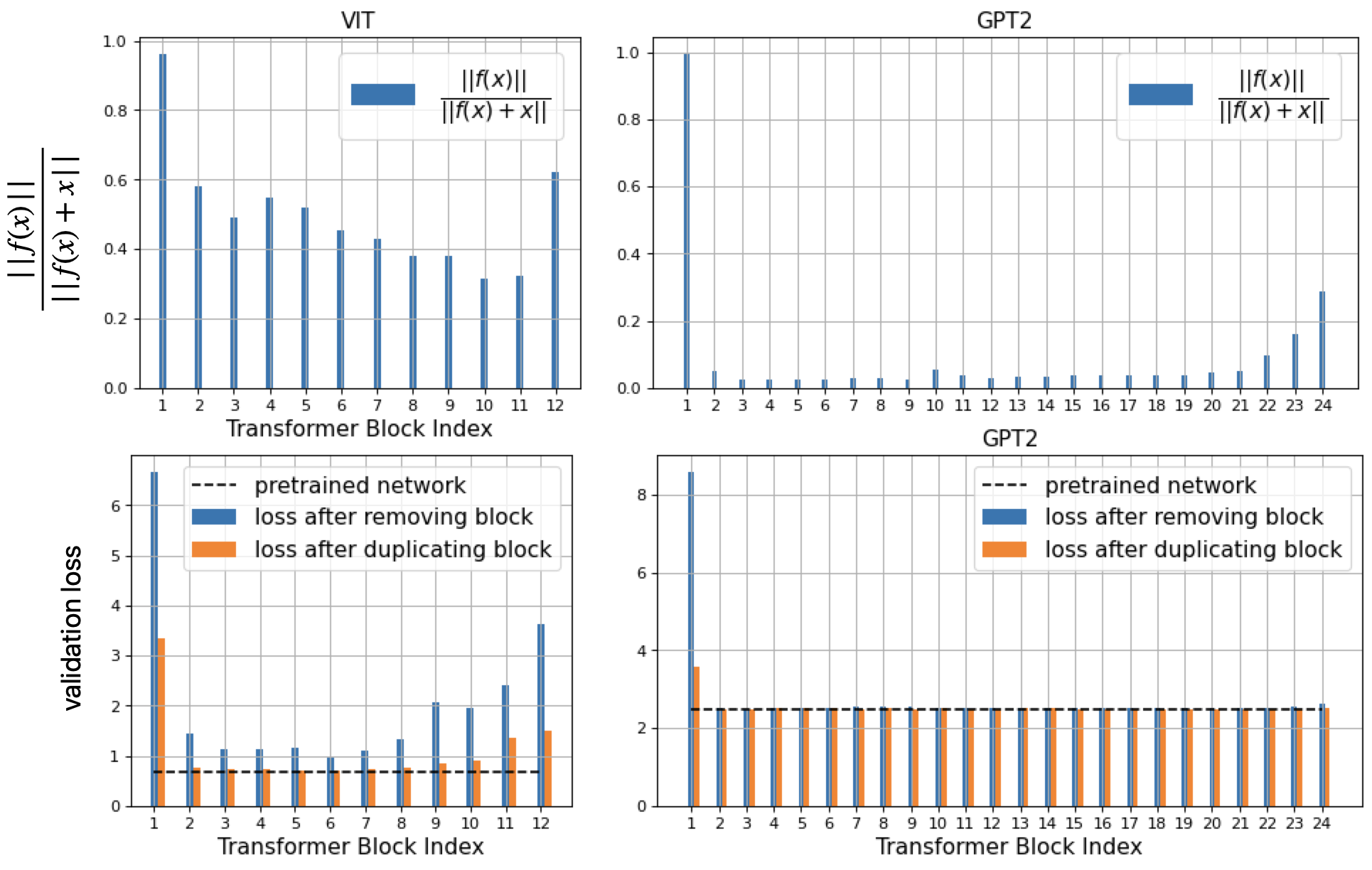}
    \caption{\textbf{Top:} Illustration of the relative magnitude of output at each transformer block, denoted as $\frac{||f(x)||}{||x+f(x)||}$. Layers with low magnitudes function similar to an identity layer, enabling their removal or duplication without substantial impact on network performance. \textbf{Bottom:} Visualization of the network's loss function as layers are removed or duplicated at various positions. Notably, as predicted by the magnitude ratio plot on the top, removing or duplicating middle layers leads to a better loss value.  }
    \label{fig:fx_remove}
\end{figure*}

In a residual transformer block, the input $x$ undergoes a transformation to produce the output $y = x + f(x)$, where $f(\cdot)$ represents the non-residual part of the block. One interpretation of the additive residual property is that the output $y$ remains similar to the input $x$, indicating that the magnitude of $f(x)$ should be small compared to $x$. In  \fig{fig:fx_remove}-Top, we present the average relative magnitude of $f(x)$ in comparison to $x+f(x)$ across image classification (VIT) and language modeling (GPT) tasks. Notably, the middle layers of the transformer networks exhibit a reduced relative magnitude in $f(x)$, indicating that these layers function similarly to identity layers.


The practical implications of this property have been studied by researchers to reduce computation complexity ~\citep{din2023jump,schwartz2020right}, to explain the model's behaviour by interpreting hidden layer representations~\cite{tenney2019bert,geva2022transformer,slobodkin2021mediators}, or to exploit the activation sparsity to improve inference speed~\citep{liu2023deja,mirzadeh2023relu}.
In this paper we study the implications of the additive residual property of transformers from a distinctive perspective: their potential to initialize a scaled-down destination model.


\noindent{\bf Changing network depth}. Since subsequent blocks of a transformer model only slightly change the hidden representation, one can either remove a single block or duplicate it without significantly altering the network's function. This technique makes it possible to create a destination network with fewer or more layers than the pretrained one. To demonstrate this effect in practice, \fig{fig:fx_remove}-bottom displays the loss of pretrained VIT and GPT transformers, when single layers are either removed or duplicated at different positions. It is worth noting that these destination networks are not fine-tuned, yet they attain meaningful loss value. Since middle layers have smaller relative value magnitude ($\frac{||f(x)||}{||x+f(x)||}$), these layers are better candidates to prune/duplicate.

\noindent{\bf Reducing embedding dimension.} To initialize a destination model with a smaller embedding dimension, we must identify neurons that have the most influence on the network's output. One way to measure this influence is by calculating the average magnitude of each neuron. Since $x$ has a significantly larger magnitude than $f(x)$ in most layers, the significance of neurons is primarily established in the early transformer blocks. In other words, if neuron number $j$ exhibits a substantial magnitude at layer $i$, it is likely that neuron $j$ also maintains a substantial magnitude in layer $i+n$. 

\begin{figure}[ht]
    \begin{subfigure}{}        \includegraphics[width=0.45\linewidth]{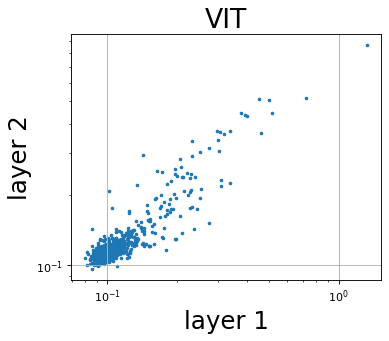}
    \end{subfigure}
    \hfill
    \begin{subfigure}{}
        \includegraphics[width=0.45\linewidth]{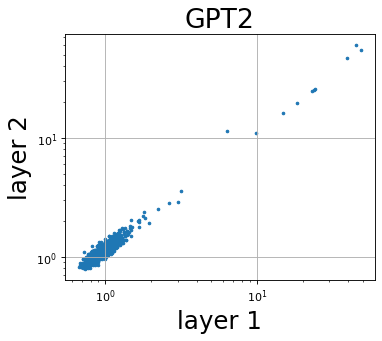}
    \end{subfigure}
    \caption{The relationship between neuron magnitudes across layers in VIT (left) and GPT2-M (right) pretrained models. Each point shows the averaged absolute value for a single neuron index. The horizontal and vertical axes represent selected layers.} \label{fig:relative_magnitudes}
\end{figure}

In \fig{fig:relative_magnitudes}, we display the averaged absolute magnitudes of neurons in paired layers of transformers trained for language modeling (GPT) and image classification (VIT). 
Our observations reveal distinct patterns. Firstly, within each layer, a subset of neurons prominently stands out with magnitudes significantly larger than the rest (please note that the figures employ a log-scale). This phenomenon implies that certain neurons within each layer exert a much more substantial influence on the network's output compared to the majority of neurons. This phenomenon has also been observed in~\citep{dettmers2022gpt3}.

Secondly, we observe a correlation between the magnitudes of neurons in different layers. Specifically, neurons with large magnitudes in layer 1 tend to exhibit similar large magnitudes in layer 2.

These observed characteristics enable us to assess the importance of neurons consistently across all transformer layers, which in turn aids in determining the most suitable weights for initializing the destination model. We will provide a systematic method to achieve this purpose in a later section of this paper.

\section{Methodology}




Suppose the parent network has $N$ blocks indexed as ${A_0, A_1, \ldots, A_{N-1}}$. Given an input $x\in\mathbb{R}^{B \times T \times d}$, where $B$ is the batch size and $T$ is the token sequence length, the transformer computes

\begin{lstlisting}
for n in range(0, N):
    x = A_n(x)
\end{lstlisting}

Our goal is to initialize a destination network with blocks ${A'_0, A'_1, \ldots, A'_{N'-1}}$, where $N' < N$. The input to the destination transformer is $x' \in \mathbb{R}^{B \times T \times d'}$, where $d' \leq d$. The destination network computes

\begin{lstlisting}
for n in range(0, N'):
    x' = A'_n(x')
\end{lstlisting}


\subsection{Subcloning Parameters} 

\begin{figure}
    \centering
    \includegraphics[width=\columnwidth]{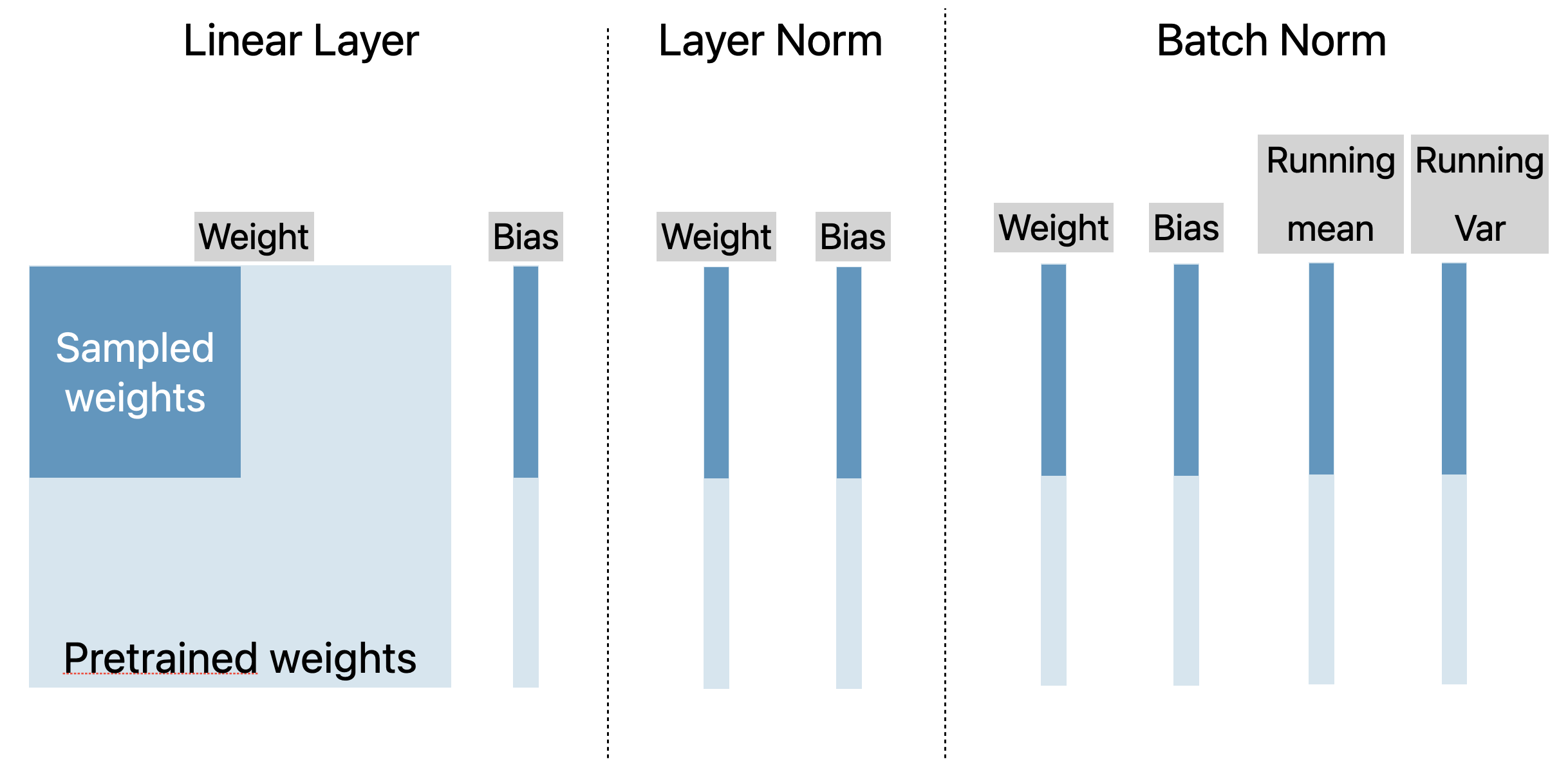}
    \caption{Sampling weights, biases, and normalization parameters. Here, the embedding dimension of the parent network is twice that of the destination network. Light blue indicates the pretrained parameters, while dark blue represents the sampled portion.}
    \label{fig:sample}
    \vspace{-0.5cm}
\end{figure}

For each block $A_n$, we need to sample a portion of its weights, biases, and normalization parameters to initialize $A'_n$. \fig{fig:sample} illustrates how the parameters are sampled. For matrix parameters in linear layers, we need to subsample both rows and columns of the pretrained matrix. For vector parameters such as  bias terms, \texttt{layer\_norm}, and \texttt{batch\_norm} parameters, we need to sample  a portion of the vector elements accordingly.

\noindent{\bf Weight re-ordering.}
To obtain the best results from weight sampling, we initially need to rearrange the rows and columns of the pretrained matrix (light blue) so that the most important weights are included in the sampled portion (dark blue). This approach allows us to select the initial rows/columns during subsampling.

\noindent{\bf Neuron importance ordering in linear layers.} A linear layer's output is a tensor of shape $B\times T \times d$, where $B$ is the batch size, $T$ is the number of tokens, and $d$ is the embedding dimension. Our goal is to rank the $d$ output neurons based on their importance. To achieve this, we run the network on a subset of data and collect the absolute values of the outputs at each transformer layer. This subset can be a small portion of the training data. In our experiments, we used 0.5\% of ImageNet for VIT and 0.003\% of Pile for GPT. 
Subsequently, we compute the average of these absolute values over the $B$ and $T$ dimensions, resulting in a $d$-dimensional vector, which we refer to as the score. 

\begin{figure}[ht]
    \begin{subfigure}{}        \includegraphics[width=0.45\linewidth]{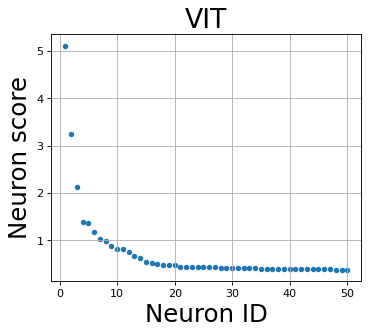}
    \end{subfigure}
    \hfill
    \begin{subfigure}{}
        \includegraphics[width=0.45\linewidth]{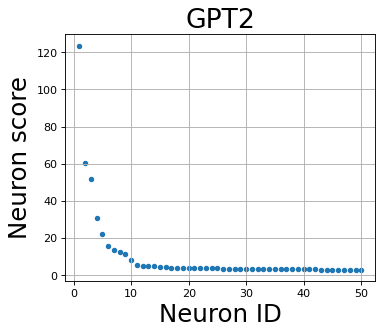}
    \end{subfigure}
    \caption{Sorted neuron scores for selected linear layers in VIT-Base (left) and GPT2-M (right). The scores are presented for the top 50 neurons with the highest scores.} \label{fig:scores_linear}
\end{figure}

In \fig{fig:scores_linear}, we provide sorted scores for example linear layers within GPT and VIT transformer blocks. Notably, certain neurons display significantly higher scores compared to others. The high-score neurons play a more substantial role in shaping the network's output. Therefore, it is essential to prioritize the inclusion of these high-scoring neurons in the destination network.

\begin{figure}[ht]
    \begin{subfigure}{}        \includegraphics[width=0.45\linewidth]{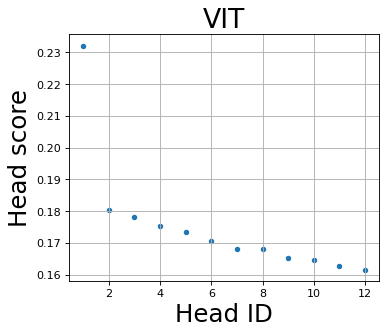}
    \end{subfigure}
    \hfill
    \begin{subfigure}{}
        \includegraphics[width=0.45\linewidth]{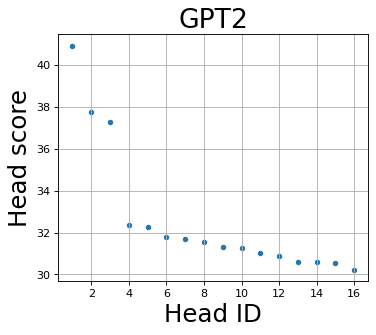}
    \end{subfigure}
    \caption{Sorted head scores for selected attention layers in VIT (left, 12 heads) and GPT-2 (right, 16 heads).} \label{fig:scores_head}
\end{figure}

\noindent{\bf Importance ordering in attention layers.} In the context of attention layers, our objective is to sort attention heads, not individual neurons. To achieve this, we begin by calculating scores for the neurons after the Softmax operation. Subsequently, we calculate the average score for each attention head by combining the scores of its constituent neurons. This process allows us to rank the attention heads based on their importance. \fig{fig:scores_head} provides a visual representation of head scores for a specific transformer layer, highlighting that certain heads exhibit significantly higher importance scores compared to others.

\begin{figure}
    \centering
    \includegraphics[width=\columnwidth]{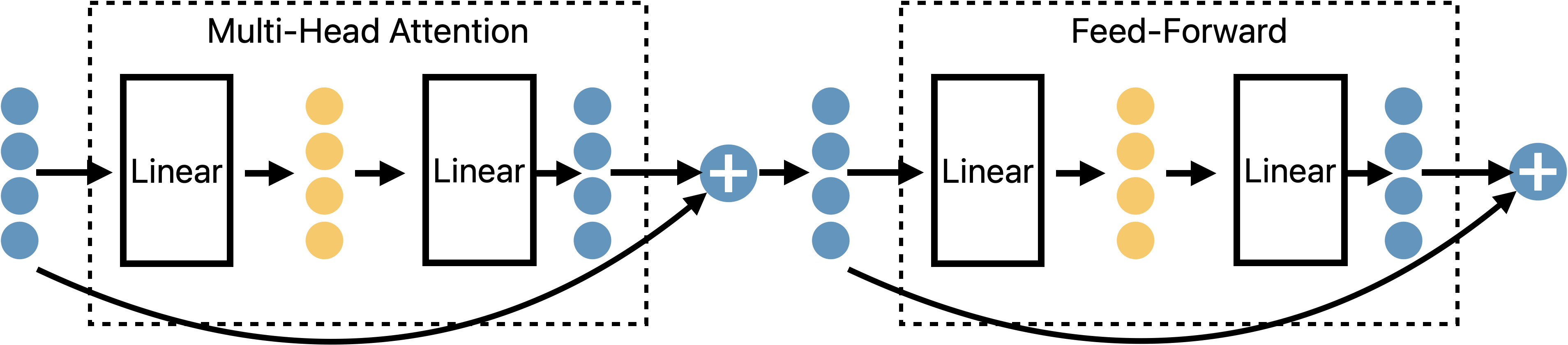}
    \caption{Block architecture in an example transformer models. Neurons connected with residual paths are illustrated with blue color. These neurons should be re-ordered consistently across all blocks to maintain the network's functionality. Activations shown in yellow are not connected through residual paths and can be re-ordered independently.}
    \label{fig:residuals}
\end{figure}

\begin{figure}[ht]
    \begin{subfigure}{}        \includegraphics[width=0.45\linewidth]{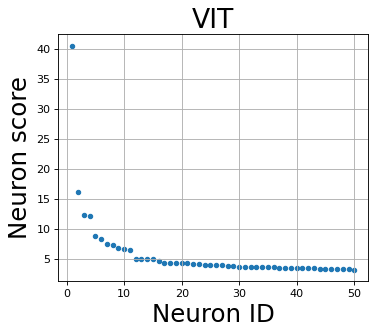}
    \end{subfigure}
    \hfill
    \begin{subfigure}{}
        \includegraphics[width=0.45\linewidth]{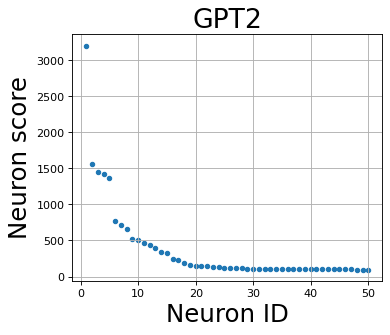}
    \end{subfigure}
    \caption{Sorted shared scores for all neurons connected by residual layers in VIT (left) and GPT-2 (right). The scores are presented for the top 50 neurons with the highest scores.} \label{fig:shared_ordering}
    \vspace{-0.5cm}
\end{figure}

\noindent{\bf Importance ordering in residual blocks.} In \fig{fig:residuals}, we illustrate a sample transformer block architecture, where activation maps are shown with either yellow (not connected through residuals) or blue (connected through residuals). When it comes to removing certain neurons from the yellow activation maps, we have the flexibility to select the removed neuron indices independently at each layer. However, with the blue activation maps, we must ensure that neurons are removed from the same positions across connected layers to maintain network connectivity. Hence, neuron re-ordering should be done persistently across layers that are connected via residual paths. To establish this shared activation order, we compute the sum of neuron importance scores across all layers connected by residual paths. In \fig{fig:shared_ordering}, we present the shared ordering of neurons across all layers in a sample transformer-based network. Notably, specific neurons consistently exhibit higher importance across all layers. This consistency can be attributed to the additive residual property of transformers, discussed in Section~\ref{sec:insights}.

\noindent{\bf Re-ordering subsequent linear layers.} 
Given the importance scores assigned to each layer's neurons in a pretrained network, we can rearrange the neurons within the layers. In the case of attention blocks, groups of neurons are reordered based on their head scores (illustrated in Figure \ref{fig:scores_head}). Neurons that do not belong to attention outputs or are not connected to residual connections are independently reordered based on their individual scores. Neurons connected through residual layers are all reordered based on the global score (depicted in Figure \ref{fig:shared_ordering}).

This reordering process is implemented by permuting the rows and columns of transformer layers. Permuting weight matrices by importance produces an equivalent model with rearranged parameters, prioritizing the most important rows and columns. From this reshuffled model, smaller networks can be subclone as illustrated earlier in \fig{fig:sample}.




\noindent{\bf Weight scaling.} When subcloning matrices from the parent to the destination network, it is crucial to maintain the standard deviation of the parent layer's neurons outputs in the destination layer. Similar methodologies have been applied to uphold the standard deviation of network layers during random initialization~\cite{glorot2010understanding}. Assuming independent and identically distributed (i.i.d.) Gaussian distributions for network weights and activations, the standard deviation of each output neuron in a linear layer is inversely proportional to the square root of the number of input neurons in that layer. Hence, to maintain the standard deviation, we multiply the subcloned weights by $\sqrt{\frac{d}{d'}}$, where $d$ and $d'$ represent the embedding sizes for the parent and destination networks, respectively. It is essential to note that scaling is unnecessary for the parameters of \texttt{layer\_norm} and \texttt{batch\_norm} layers, as well as the \texttt{bias} parameters of the linear layer.

\subsection{Subcloning Layers}
To initialize the destination network, we may remove $(N-N')$ blocks from the pretrained network so that we end up with $N'$ remaining blocks. Our experiments show that the best practice is to remove these blocks from the middle of the network, (see Figure~\ref{fig:fx_remove} and discussion in Section~\ref{sec:insights}). 

\section{Experiments}

We conducted experiments on two different data modalities.

\noindent{\bf Image Classification. } For our image classification task, we employed the Vision Transformer (VIT) models introduced in~\cite{dosovitskiy2020image}. These models were trained on the ImageNet classification dataset~\cite{deng2009imagenet}. The VIT family comprises various versions, each offering different levels of accuracy at varying computation costs. We selected VIT-B as the parent architecture, and our destination architecture was defined to have the same number of layers but with half the embedding dimension per layer. The pretrained network for our destination architecture was obtained from~\cite{faghri2023reinforce}. We conducted the training for this model on a single machine equipped with 8 NVIDIA V100 GPUs.

\noindent{\bf Language Modeling.} 
In this experiment, we focused on training GPT-2 models for next token prediction, which were originally introduced by~\cite{radford2019language}, using the Pile dataset~\cite{gao2020pile}. For our parent and destination architectures, we selected GPT-L and GPT-M, respectively. The pretrained model we utilized for this experiment was sourced from the HuggingFace repository~\cite{gpt2_hf}. In terms of architectural specifics, the parent model was comprised of 36 layers with an embedding dimension of 1280, while the destination model had 24 layers and an embedding dimension of 1024. The training process was conducted on a total of 12 machines, each of which was equipped with 8 NVIDIA V100 GPUs.

For each of the tasks mentioned above, we employed two approaches to train the destination network: random initialization and weight subcloning from the parent model. In each experiment, we fine-tuned the learning rate and weight decay parameters to ensure the fastest convergence. Additional information and specific details regarding these parameters can be found in \tab{tab:learning_rates}. We perform ablation studies for choosing these parameters in Section~\ref{sec:ablation}. Note that the curve here shows our best practice for subcloning. More alternatives and their performance will be discussed later in our ablation studies.

\begin{table*}[ht]
\centering
\caption{Training parameters. Random initialization requires large learning rate and weight decay to achieve a better accuracy, whereas weight subcloning works with small learning rate and weight decay parameters.}\label{tab:learning_rates}
\begin{tabular}{cccccc}
\multicolumn{2}{c}{Experiment}                          & Scheduler               & Optimizer              & Learning Rate & Weight Decay \\ \hline
\multirow{2}{*}{VIT (ImageNet)} & random initialization & \multirow{2}{*}{Cosine Annealing} & \multirow{2}{*}{Adam}  & 0.002         & 0.2          \\
                                & weight subcloning   &                         &                        & 0.0001        & 0.005        \\ \hline
\multirow{2}{*}{GPT2 (Pile)}    & random initialization & \multirow{2}{*}{Cosine Annealing} & \multirow{2}{*}{Adam} & 0.0001        & 0.1          \\
                                & weight subcloning   &                         &                        & 0.0001        & 0.001       
\end{tabular}
\end{table*}

\begin{figure}[ht]
    \begin{subfigure}{}
        \includegraphics[width=0.45\linewidth]{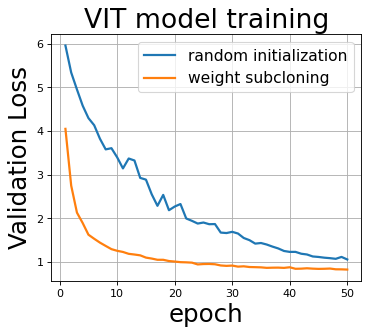}
    \end{subfigure}
    \hfill
    \begin{subfigure}{}
        \includegraphics[width=0.45\linewidth]{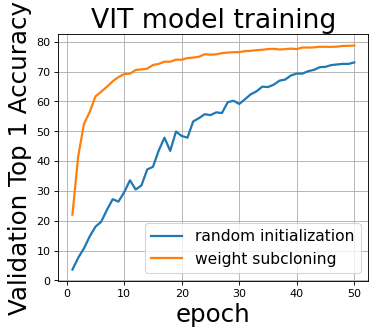}
    \end{subfigure}
    \caption{Validation loss and accuracy for VIT network trained on ImageNet for random initialization (blue) and weight subcloning (orange).} \label{fig:imagenet_training}
\end{figure}

\begin{figure}[ht]
    \begin{subfigure}{}
        \includegraphics[width=0.45\linewidth]{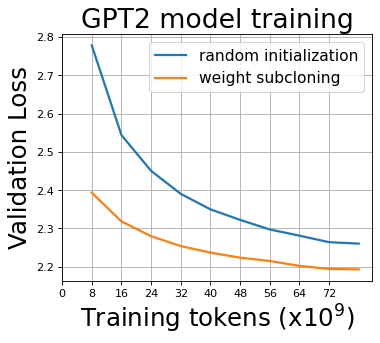}
    \end{subfigure}
    \hfill
    \begin{subfigure}{}
        \includegraphics[width=0.45\linewidth]{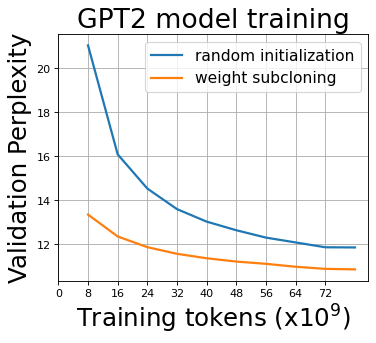}
    \end{subfigure}
    \caption{Validation loss and perplexity for GPT2 network trained on Pile for random initialization (blue) and weight subcloning (orange).} \label{fig:pile_training}
    \vspace{-0.5cm}
\end{figure}

In \fig{fig:imagenet_training}, we compare convergence curves between random initialization and weight subcloning for the ImageNet task. Additionally, for language modeling, we present the loss and perplexity in \fig{fig:pile_training}. Our experiments demonstrate the significant advantages of weight subcloning in terms of training convergence. Models initialized with weight subcloning exhibit faster convergence, achieving higher accuracy within a limited number of training epochs.

For instance, for reaching accuracy of 70\% on ImageNet, random initialization necessitates 40 epochs, while weight subcloning achieves the same accuracy in only 10 epochs, representing a $4 \times$ faster training process. In the case of GPT-2 training, random initialization requires $64\times 10^{9}$ tokens to reach a perplexity of 12, while weight subcloning accomplishes this in just $64\times 10^{9}$ tokens, again demonstrating a  $4\times$ training speedup.

\subsection{Ablation Studies}\label{sec:ablation}

\begin{figure}[ht]
    \begin{subfigure}{}
        \includegraphics[width=0.45\linewidth]{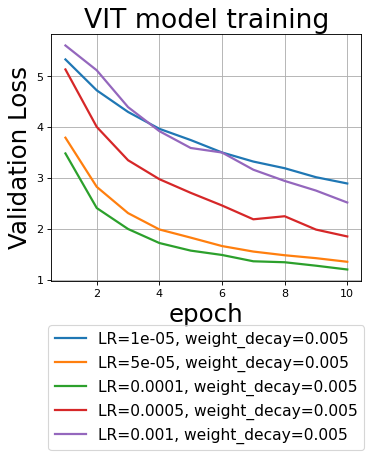}
    \end{subfigure}
    \hfill
    \begin{subfigure}{}
        \includegraphics[width=0.46\linewidth]{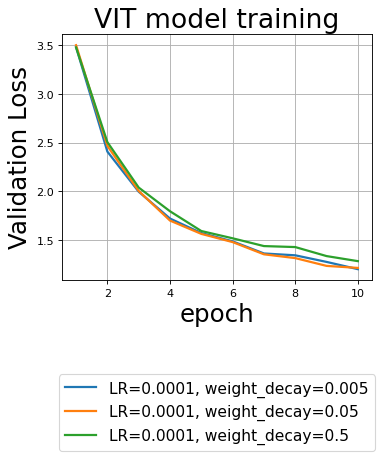}
    \end{subfigure}
    \caption{Validation loss for destination models when trained with different learning rates and weight decays. } \label{fig:learning_rate}
\end{figure}

\noindent{\bf Effect of learning rate and weight decay.} In \fig{fig:learning_rate}, we examine the convergence behavior of the VIT destination model for different learning rates and weight decays. When dealing with random initialization, it's often beneficial to employ an initialization that allows for a higher learning rate, which can facilitate improved convergence~\cite{zhuang2023survey}. However, our findings reveal a distinct pattern with weight subcloning, where lower learning rates tend to result in superior convergence. When initialized randomly, the ideal learning rate and weight decay values are 0.002 and 0.2, respectively. In contrast, when utilizing weight subcloning, the optimal values for learning rate and weight decay are 0.0001 and 0.005, respectively.

This phenomenon can be attributed to the fact that weight subcloning initializes a network that is likely to be positioned closer to a local optimum. Training such a network with a high learning rate can lead to catastrophic forgetting~\cite{french1999catastrophic}. Furthermore, models with distilled weights have already undergone weight decay and other forms of regularization during the training of the parent model. Imposing strong weight decay during the destination model training can adversely impact accuracy.

\begin{figure}[ht]
    \centering
    \includegraphics[width=0.5\linewidth]{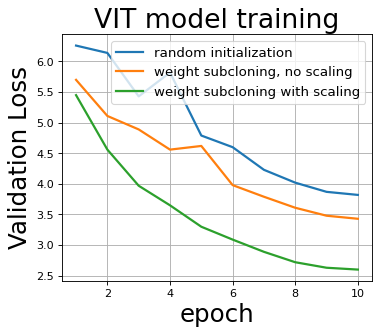}
    \caption{Validation loss for destination models with random initialization (blue), weight subcloning without scaling (orange), and weight subcloning with scaling (green).} \label{fig:scaling}
    \vspace{-0.25cm}
\end{figure}

\noindent{\bf Effect of weight scaling.} \fig{fig:scaling} illustrates the convergence rate of the destination model in three distinct scenarios: random initialization, weight subcloning without scaling, and weight subcloning with scaling. For this experiment, the parent model is VIT-H, trained on the ImageNet dataset, and the destination model shares the same number of layers but has half the embedding dimension.

The results indicate that weight scaling significantly enhances the convergence rate. This improvement is attributed to weight scaling's ability to enforce consistent standard deviations in the output of linear layers within the destination model, aligning them with the parent model and ultimately leading to more efficient convergence.

\noindent{\bf Effect of parent Model Architecture.} When selecting a parent model for training a destination model, there are often several pretrained models available, each with varying sizes and accuracies. It's common to consider choosing the parent model with the highest accuracy, assuming it would lead to better convergence during destination model training. However, our experiments reveal that opting for the highest accuracy parent model doesn't necessarily result in improved convergence for the destination model.

\begin{figure}[ht]
    \centering
    \includegraphics[width=0.5\columnwidth]{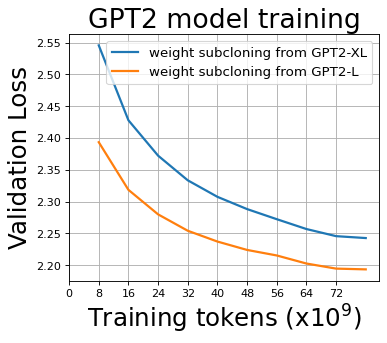}
    \caption{Validation loss for GPT2-M destination model when weights are subcloned from GPT2-L and GPT2-XL.} \label{fig:parent_effect}
\end{figure}

\fig{fig:parent_effect} demonstrates the convergence of the loss function for the destination GPT2-M model when different parent models are used for weight subcloning. In this particular experiment, the destination GPT2-M model has 24 layers with an embedding dimension of 1024. Two candidate parent models are considered: GPT2-L, which boasts 36 layers and an embedding dimension of 1280, and GPT2-XL, with 48 layers and an embedding dimension of 1600. Although GPT2-XL exhibits higher accuracy than GPT2-L, our findings suggest that GPT2-L is a more suitable choice for weight subcloning to GPT2-M. The reason behind this choice is that the architecture of GPT2-L is closer to that of GPT2-M, which leads to more effective weight subcloning and improved convergence.

\begin{figure}[ht]
    \centering
    \includegraphics[width=0.5\columnwidth]{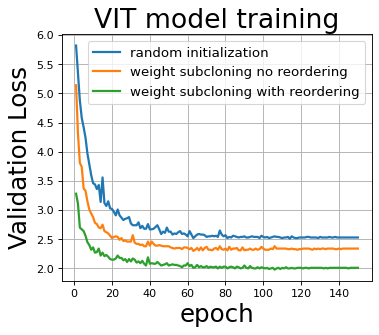}
    \caption{Validation loss for VIT destination model with random initialization (blue), weight subcloning from non-reordered parent (orange), and weight subcloning from reordered parent (green).} \label{fig:reordering_effect}
\end{figure}

\noindent{\bf Effect of neuron reordering.} \fig{fig:reordering_effect} illustrates the impact of neuron reordering on the convergence of the destination model. In this experiment, the parent model is VIT-B, while the destination model shares the same number of layers but has half the embedding size. As demonstrated in this experiment, the process of neuron reordering notably enhances the convergence of the destination network.

\section{Conclusion}
We present weight subcloning, a technique for initializing a destination transformer network using weights from a pretrained parent network. The destination network can have fewer layers or a smaller embedding dimension compared to the parent network. Our subcloning method is founded on three key components: layer removal to align the number of layers between the two networks, neuron re-ordering to match the embedding dimensions, and weight scaling to match the standard deviation of neurons.

We have demonstrated that weight subcloning significantly enhances the training speed of transformer models with virtually no additional cost, aside from a one-time initialization before training. We also conducted ablation studies to explore the impact of various factors, including learning rate, weight decay, parent network size, weight scaling, and neuron re-ordering on the training convergence of the destination network.

Lastly, the subcloning method explored in this paper is designed under the assumption that the parent and destination networks belong to the same family. While the destination network can vary from the parent in terms of the number of blocks and/or the embedding dimension per layer, our study has not delved into the impact of more extensive architectural changes. These changes might include modifications to residual connections, nonlinear activations, block structures, and similar aspects. These areas remain promising topics for future research.

\bibliography{ref}

\begin{thebibliography}{33}
\providecommand{\natexlab}[1]{#1}
\providecommand{\url}[1]{\texttt{#1}}
\expandafter\ifx\csname urlstyle\endcsname\relax
  \providecommand{\doi}[1]{doi: #1}\else
  \providecommand{\doi}{doi: \begingroup \urlstyle{rm}\Url}\fi

\bibitem[Blalock et~al.(2020)Blalock, Gonzalez~Ortiz, Frankle, and Guttag]{blalock2020state}
Blalock, D., Gonzalez~Ortiz, J.~J., Frankle, J., and Guttag, J.
\newblock What is the state of neural network pruning?
\newblock \emph{Proceedings of machine learning and systems}, 2:\penalty0 129--146, 2020.

\bibitem[Cai et~al.(2019)Cai, Gan, Wang, Zhang, and Han]{cai2019once}
Cai, H., Gan, C., Wang, T., Zhang, Z., and Han, S.
\newblock Once-for-all: Train one network and specialize it for efficient deployment.
\newblock \emph{arXiv preprint arXiv:1908.09791}, 2019.

\bibitem[Dai et~al.(2019)Dai, Yang, Yang, Carbonell, Le, and Salakhutdinov]{dai2019transformer}
Dai, Z., Yang, Z., Yang, Y., Carbonell, J., Le, Q.~V., and Salakhutdinov, R.
\newblock Transformer-xl: Attentive language models beyond a fixed-length context.
\newblock \emph{arXiv preprint arXiv:1901.02860}, 2019.

\bibitem[Deng et~al.(2009)Deng, Dong, Socher, Li, Li, and Fei-Fei]{deng2009imagenet}
Deng, J., Dong, W., Socher, R., Li, L.-J., Li, K., and Fei-Fei, L.
\newblock Imagenet: A large-scale hierarchical image database.
\newblock In \emph{2009 IEEE conference on computer vision and pattern recognition}, pp.\  248--255. Ieee, 2009.

\bibitem[Dettmers et~al.(2022)Dettmers, Lewis, Belkada, and Zettlemoyer]{dettmers2022gpt3}
Dettmers, T., Lewis, M., Belkada, Y., and Zettlemoyer, L.
\newblock Gpt3. int8 (): 8-bit matrix multiplication for transformers at scale.
\newblock \emph{Advances in Neural Information Processing Systems}, 35:\penalty0 30318--30332, 2022.

\bibitem[Din et~al.(2023)Din, Karidi, Choshen, and Geva]{din2023jump}
Din, A.~Y., Karidi, T., Choshen, L., and Geva, M.
\newblock Jump to conclusions: Short-cutting transformers with linear transformations.
\newblock \emph{arXiv preprint arXiv:2303.09435}, 2023.

\bibitem[Dosovitskiy et~al.(2020)Dosovitskiy, Beyer, Kolesnikov, Weissenborn, Zhai, Unterthiner, Dehghani, Minderer, Heigold, Gelly, et~al.]{dosovitskiy2020image}
Dosovitskiy, A., Beyer, L., Kolesnikov, A., Weissenborn, D., Zhai, X., Unterthiner, T., Dehghani, M., Minderer, M., Heigold, G., Gelly, S., et~al.
\newblock An image is worth 16x16 words: Transformers for image recognition at scale.
\newblock \emph{arXiv preprint arXiv:2010.11929}, 2020.

\bibitem[Faghri et~al.(2023)Faghri, Pouransari, Mehta, Farajtabar, Farhadi, Rastegari, and Tuzel]{faghri2023reinforce}
Faghri, F., Pouransari, H., Mehta, S., Farajtabar, M., Farhadi, A., Rastegari, M., and Tuzel, O.
\newblock Reinforce data, multiply impact: Improved model accuracy and robustness with dataset reinforcement.
\newblock \emph{arXiv preprint arXiv:2303.08983}, 2023.

\bibitem[Frankle \& Carbin(2018)Frankle and Carbin]{frankle2018lottery}
Frankle, J. and Carbin, M.
\newblock The lottery ticket hypothesis: Finding sparse, trainable neural networks.
\newblock \emph{arXiv preprint arXiv:1803.03635}, 2018.

\bibitem[French(1999)]{french1999catastrophic}
French, R.~M.
\newblock Catastrophic forgetting in connectionist networks.
\newblock \emph{Trends in cognitive sciences}, 3\penalty0 (4):\penalty0 128--135, 1999.

\bibitem[Gao et~al.(2020)Gao, Biderman, Black, Golding, Hoppe, Foster, Phang, He, Thite, Nabeshima, et~al.]{gao2020pile}
Gao, L., Biderman, S., Black, S., Golding, L., Hoppe, T., Foster, C., Phang, J., He, H., Thite, A., Nabeshima, N., et~al.
\newblock The pile: An 800gb dataset of diverse text for language modeling.
\newblock \emph{arXiv preprint arXiv:2101.00027}, 2020.

\bibitem[Geva et~al.(2022)Geva, Caciularu, Wang, and Goldberg]{geva2022transformer}
Geva, M., Caciularu, A., Wang, K.~R., and Goldberg, Y.
\newblock Transformer feed-forward layers build predictions by promoting concepts in the vocabulary space.
\newblock \emph{arXiv preprint arXiv:2203.14680}, 2022.

\bibitem[Glorot \& Bengio(2010)Glorot and Bengio]{glorot2010understanding}
Glorot, X. and Bengio, Y.
\newblock Understanding the difficulty of training deep feedforward neural networks.
\newblock In \emph{Proceedings of the thirteenth international conference on artificial intelligence and statistics}, pp.\  249--256. JMLR Workshop and Conference Proceedings, 2010.

\bibitem[Gou et~al.(2021)Gou, Yu, Maybank, and Tao]{gou2021knowledge}
Gou, J., Yu, B., Maybank, S.~J., and Tao, D.
\newblock Knowledge distillation: A survey.
\newblock \emph{International Journal of Computer Vision}, 129:\penalty0 1789--1819, 2021.

\bibitem[Han et~al.(2022)Han, Wang, Chen, Chen, Guo, Liu, Tang, Xiao, Xu, Xu, et~al.]{han2022survey}
Han, K., Wang, Y., Chen, H., Chen, X., Guo, J., Liu, Z., Tang, Y., Xiao, A., Xu, C., Xu, Y., et~al.
\newblock A survey on vision transformer.
\newblock \emph{IEEE transactions on pattern analysis and machine intelligence}, 45\penalty0 (1):\penalty0 87--110, 2022.

\bibitem[Han et~al.(2015)Han, Pool, Tran, and Dally]{han2015learning}
Han, S., Pool, J., Tran, J., and Dally, W.
\newblock Learning both weights and connections for efficient neural network.
\newblock \emph{Advances in neural information processing systems}, 28, 2015.

\bibitem[He et~al.(2016)He, Zhang, Ren, and Sun]{he2016deep}
He, K., Zhang, X., Ren, S., and Sun, J.
\newblock Deep residual learning for image recognition.
\newblock In \emph{Proceedings of the IEEE conference on computer vision and pattern recognition}, pp.\  770--778, 2016.

\bibitem[He et~al.(2017)He, Zhang, and Sun]{he2017channel}
He, Y., Zhang, X., and Sun, J.
\newblock Channel pruning for accelerating very deep neural networks.
\newblock In \emph{Proceedings of the IEEE international conference on computer vision}, pp.\  1389--1397, 2017.

\bibitem[HuggingFace(2023)]{gpt2_hf}
HuggingFace.
\newblock Name of the model checkpoint.
\newblock \url{https://huggingface.co/gpt2-large}, 2023.
\newblock Hugging Face model checkpoint. Accecced on June 2023.

\bibitem[Lin et~al.(2022)Lin, Xie, Wang, Yu, Chang, Liang, and Wang]{lin2022knowledge}
Lin, S., Xie, H., Wang, B., Yu, K., Chang, X., Liang, X., and Wang, G.
\newblock Knowledge distillation via the target-aware transformer.
\newblock In \emph{Proceedings of the IEEE/CVF Conference on Computer Vision and Pattern Recognition}, pp.\  10915--10924, 2022.

\bibitem[Lin et~al.(2020)Lin, Li, Wang, Li, Du, Xiao, and Zhu]{lin2020weight}
Lin, Y., Li, Y., Wang, Z., Li, B., Du, Q., Xiao, T., and Zhu, J.
\newblock Weight distillation: Transferring the knowledge in neural network parameters.
\newblock \emph{arXiv preprint arXiv:2009.09152}, 2020.

\bibitem[Liu et~al.(2023)Liu, Wang, Dao, Zhou, Yuan, Song, Shrivastava, Zhang, Tian, Re, et~al.]{liu2023deja}
Liu, Z., Wang, J., Dao, T., Zhou, T., Yuan, B., Song, Z., Shrivastava, A., Zhang, C., Tian, Y., Re, C., et~al.
\newblock Deja vu: Contextual sparsity for efficient llms at inference time.
\newblock In \emph{International Conference on Machine Learning}, pp.\  22137--22176. PMLR, 2023.

\bibitem[Mirzadeh et~al.(2023)Mirzadeh, Alizadeh, Mehta, Mundo, Tuzel, Samei, Rastegari, and Farajtabar]{mirzadeh2023relu}
Mirzadeh, I., Alizadeh, K., Mehta, S., Mundo, C. C.~D., Tuzel, O., Samei, G., Rastegari, M., and Farajtabar, M.
\newblock Relu strikes back: Exploiting activation sparsity in large language models, 2023.

\bibitem[Park et~al.(2022)Park, Kim, Oh, Seo, Lee, Kim, Moon, Lim, Park, and Ye]{park2022self}
Park, S., Kim, G., Oh, Y., Seo, J.~B., Lee, S.~M., Kim, J.~H., Moon, S., Lim, J.-K., Park, C.~M., and Ye, J.~C.
\newblock Self-evolving vision transformer for chest x-ray diagnosis through knowledge distillation.
\newblock \emph{Nature communications}, 13\penalty0 (1):\penalty0 3848, 2022.

\bibitem[Radford et~al.(2019)Radford, Wu, Child, Luan, Amodei, Sutskever, et~al.]{radford2019language}
Radford, A., Wu, J., Child, R., Luan, D., Amodei, D., Sutskever, I., et~al.
\newblock Language models are unsupervised multitask learners.
\newblock \emph{OpenAI blog}, 1\penalty0 (8):\penalty0 9, 2019.

\bibitem[Schwartz et~al.(2020)Schwartz, Stanovsky, Swayamdipta, Dodge, and Smith]{schwartz2020right}
Schwartz, R., Stanovsky, G., Swayamdipta, S., Dodge, J., and Smith, N.~A.
\newblock The right tool for the job: Matching model and instance complexities.
\newblock \emph{arXiv preprint arXiv:2004.07453}, 2020.

\bibitem[Slobodkin et~al.(2021)Slobodkin, Choshen, and Abend]{slobodkin2021mediators}
Slobodkin, A., Choshen, L., and Abend, O.
\newblock Mediators in determining what processing bert performs first.
\newblock \emph{arXiv preprint arXiv:2104.06400}, 2021.

\bibitem[Tenney et~al.(2019)Tenney, Das, and Pavlick]{tenney2019bert}
Tenney, I., Das, D., and Pavlick, E.
\newblock Bert rediscovers the classical nlp pipeline.
\newblock \emph{arXiv preprint arXiv:1905.05950}, 2019.

\bibitem[Wang et~al.(2021{\natexlab{a}})Wang, Gong, Li, Liu, and Chandra]{wang2021alphanet}
Wang, D., Gong, C., Li, M., Liu, Q., and Chandra, V.
\newblock Alphanet: Improved training of supernets with alpha-divergence.
\newblock In \emph{International Conference on Machine Learning}, pp.\  10760--10771. PMLR, 2021{\natexlab{a}}.

\bibitem[Wang et~al.(2021{\natexlab{b}})Wang, Li, Gong, and Chandra]{wang2021attentivenas}
Wang, D., Li, M., Gong, C., and Chandra, V.
\newblock Attentivenas: Improving neural architecture search via attentive sampling.
\newblock In \emph{Proceedings of the IEEE/CVF conference on computer vision and pattern recognition}, pp.\  6418--6427, 2021{\natexlab{b}}.

\bibitem[Yu et~al.(2020)Yu, Jin, Liu, Bender, Kindermans, Tan, Huang, Song, Pang, and Le]{yu2020bignas}
Yu, J., Jin, P., Liu, H., Bender, G., Kindermans, P.-J., Tan, M., Huang, T., Song, X., Pang, R., and Le, Q.
\newblock Bignas: Scaling up neural architecture search with big single-stage models.
\newblock In \emph{Computer Vision--ECCV 2020: 16th European Conference, Glasgow, UK, August 23--28, 2020, Proceedings, Part VII 16}, pp.\  702--717. Springer, 2020.

\bibitem[Zhang et~al.(2023)Zhang, Song, Li, Zhou, and Song]{zhang2023survey}
Zhang, H., Song, H., Li, S., Zhou, M., and Song, D.
\newblock A survey of controllable text generation using transformer-based pre-trained language models.
\newblock \emph{ACM Computing Surveys}, 56\penalty0 (3):\penalty0 1--37, 2023.

\bibitem[Zhuang et~al.(2023)Zhuang, Liu, Pan, He, Weng, and Shen]{zhuang2023survey}
Zhuang, B., Liu, J., Pan, Z., He, H., Weng, Y., and Shen, C.
\newblock A survey on efficient training of transformers.
\newblock \emph{arXiv preprint arXiv:2302.01107}, 2023.

\end{thebibliography}
\bibliographystyle{icml2023}



\end{document}